\documentclass[runningheads]{llncs}

\usepackage{eccv}

\usepackage{eccvabbrv}

\usepackage{graphicx}
\usepackage{booktabs}
\usepackage{todonotes}
\usepackage{url}

\usepackage[accsupp]{axessibility}  % Improves PDF readability for those with disabilities.

\usepackage{hyperref}

\usepackage{orcidlink}
\usepackage{multirow}

\newsavebox\CBox
\def\textBF#1{\sbox\CBox{#1}\resizebox{\wd\CBox}{\ht\CBox}{\textbf{#1}}}

\begin{document}

\title{Automatic Die Studies for Ancient Numismatics} 

\titlerunning{Automatic Die Studies}

\author{Clément Cornet\inst{1}\orcidlink{0009-0001-5829-9741} \and
Héloïse Aumaître\inst{2}\orcidlink{0009-0003-5850-5087} \and
Romaric Besançon\inst{1}\orcidlink{0000-0003-1331-5768} \and
Julien Olivier\inst{3}\orcidlink{0000-0003-1579-1311} \and
Thomas Faucher\inst{1}\orcidlink{0000-0003-1016-3340} \and
Hervé Le Borgne\inst{2}\orcidlink{0000-0003-0520-8436}}

\authorrunning{C. Cornet et al.}

\institute{Université Paris-Saclay, CEA, List, F-91120, Palaiseau, France\\
\email{clement.cornet/romaric.besancon/herve.le-borgne@cea.fr} \and
Centre d’Études Alexandrines (UAR 3134 CNRS/Ifao), Alexandria, Egypt\\
\email{heloise.aumaitre/thomas.faucher@cea.com.eg} \and
IRAMAT-CEB (UMR 5060, CNRS-Université d'Orléans), France\\
\email{julien.olivier@bnf.fr}
}

\maketitle

\begin{abstract}
Die studies are fundamental to quantifying ancient monetary production, providing insights into the relationship between coinage, politics, and history. The process requires tedious manual work, which limits the size of the corpora that can be studied. Few works have attempted to automate this task, and none have been properly released and evaluated from a computer vision perspective. We propose a fully automatic approach that introduces several innovations compared to previous methods. We rely on fast and robust local descriptors matching that is set automatically. Second, the core of our proposal is a clustering-based approach that uses an intrinsic metric (that does not need the ground truth labels) to determine its critical hyper-parameters. We validate the approach on two corpora of Greek coins, propose an automatic implementation and evaluation of previous baselines, and show that our approach significantly outperforms them.
\end{abstract}

\section{Introduction}
\label{sec:intro}
The quantification of ancient monetary production helps understand the mint outputs and the relationship between coinage, politics and history~\cite{bland2018quantifying}. This quantification primarily relies on die studies, which compare each coin in a series (with the same types) to determine the number of engraved tools (the dies) used to strike coins. Using statistical models~\cite{carter1983die_link_stat, esty2011geometrical_model}, it is then possible to estimate the number of dies used for each series and, consequently, the volume of coin produced. However, comparing each coin to the others in a corpus is a long and tedious manual work, which grows quadratically $\mathcal{O}(N^2)$ with the size $N$ of the corpus, limiting current dies studies to a few thousand coins. An automatic approach to conducting die studies would not only offer an appreciable gain of time but  also open up prospects for larger-scale studies.

\begin{figure}
     \centering
     \begin{subfigure}[b]{0.3\textwidth}
         \centering
         \includegraphics[width=\textwidth]{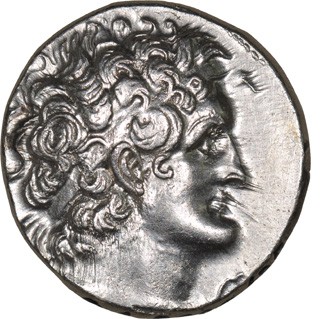}
         \caption{Die 24}
         \label{subfig:coin-d24-a}
     \end{subfigure}
     \hfill
     \begin{subfigure}[b]{0.3\textwidth}
         \centering
         \includegraphics[width=\textwidth]{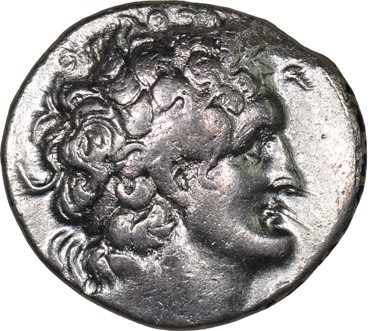}
         \caption{Die 24}
         \label{subfig:coin-d24-b}
     \end{subfigure}
     \hfill
     \begin{subfigure}[b]{0.3\textwidth}
         \centering
         \includegraphics[width=\textwidth]{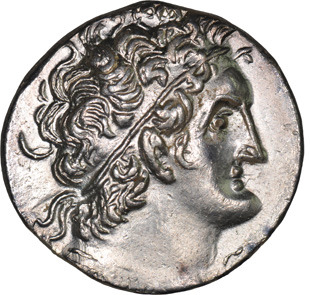}
         \caption{Die 52}
         \label{subfig:coin-d52-c}
     \end{subfigure}
        \caption{Differences between coins from different dies, Paphos \cite{faucher2017paphos}. Coins 27 (\ref{subfig:coin-d24-a}) and 28 (\ref{subfig:coin-d24-b}) were struck by the same die, while coin 95 (\ref{subfig:coin-d52-c}) was struck from a different one.}
        \label{fig:different-dies-paphos}
\end{figure}

The first attempt in that direction is CADS~\cite{taylor2020computer}, a semi-automatic approach which consists in extracting ORB local descriptors at keypoints~\cite{rublee2011orb} in each image, deriving a similarity matrix then performing hierarchical clustering to identify some groups that may correspond to the coins struck by the same die. However, the approach requires two manual operations, for filtering the keypoints and for determining the threshold used to form the final groups. Another promising approach for automated die studies is Riedones3D \cite{horache2021riedones3d} but it requires 3D-scanning every coin beforehand. The need for a 3D-Scanner is self-limiting, as it would in practice be very tedious to deploy at large scale for the millions of coins available.

In this paper, we propose a fully automated method for conducting die studies on low-resolution RGB images. Our approach involves extracting robust and fast deep learning features from coins, which are then filtered using a Random Sample Consensus method that operates without requiring hyperparameter settings. Subsequently, the matches between coins are used to construct a graph, and the coins are clustered using a graph clustering algorithm. To determine the crucial hyperparameter for graph clustering, we propose to rely on the intrinsic structural consistency of potential partitions. As a result, this method operates entirely without human intervention.

Our approach is evaluated on two die studies and compared to an automated version of CADS. It exhibits significantly better performance, while remaining fast: a die study on thousands of coins can be achieved in a couple of hours, while it takes several weeks for historians of antiquity. %ancient historians.
The code and models are made publicly available\footnote{project page: \href{https://cea-list-lasti.github.io/projects/studies/studies.html}{https://cea-list-lasti.github.io/projects/studies/studies.html}}, not only for the sake of reproducibility, but also to be used by numismatists for their future die studies.

\section{Related Works}

\paragraph{Automatic and semi-automatic Die Studies}
While most automated coin analysis systems are focused on coin types recognition \cite{manzoor2022ancient, cooper2020learning} and corpus analysis \cite{deligio2024supporting}, some works tackle the problem of automating die studies. The seminal contribution in this direction is CADS \cite{taylor2020computer}. It consists in extracting keypoints descriptors for every coin using ORB \cite{rublee2011orb} and finding the corresponding matches. From these, they construct a pairwise dissimilarity measure upon the descriptors, and perform an agglomerative clustering algorithm \cite{kaufman1990agglomerative} using this measure. The clusters obtained correspond to the coins struck by a same die. However, CADS is semi-automatic, as it requires human intervention at preprocessing and clustering stages.
In a preprint~\cite{heinecke2021unsupervised}, Heinecke \etal propose an automated (``unsupervised'') approach based on a similar pipeline as CADS, with Gaussian process keypoint extraction and a Bayesian distance microclustering algorithm~\cite{natarajan2024cohesion}. However, their code is only partially released, which makes it hard to reproduce their results. Our approach differs from both these works, with the use of more efficient deep learning local features and a fully automatic clustering method. 

The Riedones3D approach \cite{horache2021riedones3d} exhibits promising results for automated die studies, relying on 3D-scanning every coin of the corpus. It is nevertheless limited in practice since it requires to re-scan every corpus with 3D-scan, while our approach relies on the RGB images of the coins, which are easier to acquire and generally already produced by numismatists for their die studies. %currently already available for more than $2,700$ corpus of the Die Studies Database\footnote{The DSD aims to make available all the produced die-studies for the Graeco-Roman world \url{https://silver.knowledge.wiki/SILVER}}.
%%% Thomas : Die study available but not the corpus itself for each study

\paragraph{Coin recognition}
Coin image classification and recognition is a quite well-established task in computer vision, historically addressed with traditional segmentation approaches~\cite{zambanini2008segmentation, say2014segmentation_coin}. More recently, Joshi \etal proposed a hybrid approach combining SVM and neural networks to identify small parts of manufactured objects and applied it to Indian coin recognition~\cite{joshi2020flexible}. The neural network is nevertheless a classical feed-forward network that processes handmade visual feature vectors, as an alternative to the SVM. In the same vein Fonov and Ksenia also used a two-layer feed-forward neural network with visual features designed by hand such as filtered version of the coin image~\cite{fonov2021nn_coin_recognition}. A more elaborated approach was proposed~\cite{answar2021deep_ancient} which relies on two convolutional networks that directly learn the features from the image in accordance with the deep learning principles. They proposed to pool the features extracted by both neural networks with a bilinear pooling layer and added an attention layer to emphasize the salient features. This method has been applied to an image dataset of the Roman Republican coins that contains more than 18,000 images and obtained good performances of coin recognition. More generally, computer tools can be used to assist numismatists in the processing and analysis of large corpora of coins semi-automatically~\cite{deligio2024supporting}. Recognizing a coin image nevertheless differs from the task we address in this paper, since the recognition of the die that is common to several coins requires distinguishing finer details. 

\paragraph{Image Matching}

The goal of image matching is to find matching keypoints from a pair of images of the same instance (object or building) taken with different conditions (luminosity, orientation...). Traditional image matching techniques use classic keypoints detectors \cite{harris1988combined} with independent descriptors \cite{rublee2011orb,lowe2004distinctive}. The lack of link between keypoints detection and descriptions leading to limited performance, more recent approaches adopted a joint detection and description \cite{tyszkiewicz2020disk}. Learned feature matchers \cite{lindenberger2023lightglue,jiang2024omniglue,edstedt2024roma} have shown remarkable improvements both in terms of accuracy and robustness. In addition, some recent techniques have shown promising results by performing joint segmentation and matching \cite{zhang2024mesa}. Lately, complex network architectures lead to a massive increase in runtime, yet speed becomes an important objective for modern approaches \cite{potje2024xfeat}. 

However, the task we address differs from the image matching one. While the latter aims at finding the matches between two images of the same instance, we consider a pair of images of two different instances of coins. We aim at identifying latent common properties, namely the fact that they were struck by the same die. Although the point of view and the illuminating conditions are usually more regular than in traditional image matching tasks, the two instances differ due to differentiated wear and tear of both coins and dies.

%%%=============================================================
\section{Method}
\begin{figure}[!tb]
    \centering
    \includegraphics[width=\textwidth]{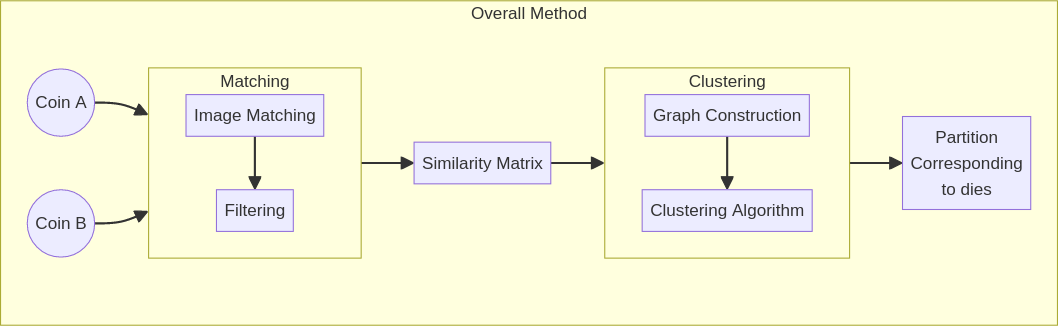}
    \caption{Overview of the method}
    \label{fig:method-overview}
\end{figure}

We adopt a general pipeline similar to the one proposed in CADS, that comprises two main stages (see Figure \ref{fig:method-overview}). First, we extract matches from every pair of images from a collection in order to build a pairwise similarity matrix. Then, we perform a clustering using this similarity matrix to get a partition corresponding to the dies.

\subsection{Coin (dis)similarities}
Similarly to the way numismatists compare coins in die studies, we look for matching features from every pair of coins. Differing from previous works \cite{taylor2020computer,heinecke2021unsupervised}, we use deep-learning matches, namely XFeat \cite{potje2024xfeat}. XFeat is an image matching technique that obtains remarkable results on image matching benchmarks \cite{mishkin2015wxbs, li2018megadepth}, while being extremely fast, yet able to satisfy the need of time-gain required for larger-scale die-studies. An important hyperparameter is $top_k$, the maximum number of regions to consider for matching. As XFeat is trained on $SfM$ tasks such as relative camera pose estimation, retraining it on coins (all taken from the same position above coins) would not make sense. We thus use pretrained weights, learnt from a blend of Megadepth \cite{li2018megadepth} and synthetically warped COCO \cite{lin2014microsoft}.

Most image matching techniques - including XFeat - are designed to find matching keypoints between two images from a same object or building, while being robust to changes in orientation or luminosity. Here, we analyze different coin instances, with the hypothesis that coins from the same die must have similar structures. We expect the keypoints detector to be robust with regard to the preservation state of coins, die wear or the way coins were struck. 

In some cases, in particular when matching images from different coin instances, we could have irrelevant matches: matches sometimes occur in the background of the image, or on wear traces, mostly on the edge of coins. CADS~\cite{taylor2020computer} tackles this problem by filtering out the edges of the coins. However, this approach requires a human intervention to select the coin region to consider, which is not viable on large datasets. Moreover, important keypoints could still be near the edges of a coin. Totally cutting them off could lead to a performance loss.

We use the RANSAC-based MAGSAC++~\cite{barath2020magsac++} for filtering inlier matches from outliers. Assuming noise $\sigma$ follows an uniform distribution, it fits an homography estimation via Iteratively Reweighted Least Square. At each step, the weights for each sample are recomputed from the marginalized over $\sigma$ likelihood of a sample being an inlier ($\sigma$-consensus). Once the model has converged, we keep the matches that are identified as inliers. MAGSAC++ is robust, and does not require a user-defined threshold over $\sigma$ to consider a sample as an outlier.

The number of filtered matches between two coins can then be interpreted and handled as a similarity measure. Then, repeating the process for every pair of coins in a collection produces a pairwise similarity matrix $M$. We derive a pairwise dissimilarity matrix as $D=\max(M)-M$, better suited for further analysis. Other decreasing functions were tested to derive $D$ from $M$, without noting significant difference in our experimental results. 

\subsection{Adaptive Graph Label Propagation}\label{sec:clustering}
We construct a graph $\mathcal{G}$ which nodes correspond to each coin of the considered corpus. Two nodes are connected if the number of matches between corresponding coins exceeds a threshold $\tau$. When $\tau$ is chosen appropriately, coins should be connected only if they were struck by the same die. This approach of using an unweighted graph aligns more closely with traditional die studies.

To obtain our final partition, we perform graph clustering (or community extraction) over $\mathcal{G}$, using the Label Propagation Clustering algorithm \cite{raghavan2007near}. This approach first assigns a unique label to each node. It then iteratively updates each node's label to the most common label among its neighbors, processing nodes in a random order until convergence.

To determine the optimal threshold $\tau^*$, we compute the partitions corresponding to every possible value of $\tau$. For each of them, we compute the Silhouette Coefficient \cite{rousseeuw1987silhouettes} that does not require ground truth clustering and only evaluates the coherence of the inner structure of a partition, given a dissimilarity measure. For a sample $x_i$ that is in the $k^{th}$ cluster $C_k$, it assesses whether it is closer to the other samples of its own cluster, or to the next-nearest cluster. We consider the value
\begin{equation}
    s(x_i) = \frac{b(x_i) - a(x_i)}{\max(a(x_i), b(x_i))}
\end{equation}

With $a(x_i)=\frac{1}{|C_k|-1}\sum_{x_j\in C_k, x_j\ne x_i}D(x_i,x_j)$ the average distance to same-cluster samples, and $b(x_i)=\min_{l\ne k}\frac{1}{C_l}\sum_{j\in C_l}D(x_i,x_j)$ the average distance to next-nearest cluster samples. The Silhouette Coefficient of a whole dataset is then the average of $s(.)$ for all samples in the dataset. %To evaluate the Silhouette from the number of matches $M$, we first compute the dissimilarity $D = max(M) - M$, before computing the Silhouette of $D$. 

The Silhouette Coefficient ranges from -1 to 1, with scores around zero for overlapping clusters, and close to 1 for dense and separated clusters, which are properties that the optimal partition should satisfy. Finally, we keep the partition that has the highest Silhouette Coefficient. The resulting Adaptive Graph Label Propagation (AGLP) approach thus allows to determine the groups that correspond to each die without any manual hyperparameter setting.

\section{Experiments}
\subsection{Baseline: a fully automatic CADS approach}
In order to compare our approach to previous works, we reproduce the CADS model \cite{taylor2020computer}, with some adaptations to make the model fully automated, since the original model requires manual operations:

\begin{itemize}
    \item CADS only considers matches within a user-defined circle, in order to filter out edges. Human intervention is needed because the optimal radius of the circle varies from one coin to another. Since this process is not suited for automation, we ignore this filtering step in our fully automatic version of the model;
    \item CADS builds a hierarchy between coins with agglomerative nesting \cite{kaufman1990agglomerative}, but does not automatically decide where to cut it to build a partition, leaving the decision to the user.
\end{itemize}
We propose a first baseline $CADS-AG^*$ that reproduces the semi-automatic approach CADS with a human user able to find the best possible cut for the automatic clustering. The second baseline is actually automatic, using a HDBSCAN clustering from the pairwise distance matrix, with a minimum cluster size of 2 ($CADS-HDBSCAN$).

\subsection{Evaluation framework}
\subsubsection{Datasets}
We perform the evaluation on two collections of coins, with their associated die analysis : the Paphos \cite{faucher2017paphos} and Tanis 1986 \cite{faucher2017tanis86} collections, with respectively 2484 and 295 coins. We consider the coin images at a resolution of $288 \times 288$ pixels for Paphos, and $480 \times 512$ for Tanis 1986. In \cite{taylor2020computer}, the CADS model was evaluated on a subset of 200 coins of the Paphos collection.

\begin{figure}[!tb]
    \centering
    \includegraphics[scale=0.4]{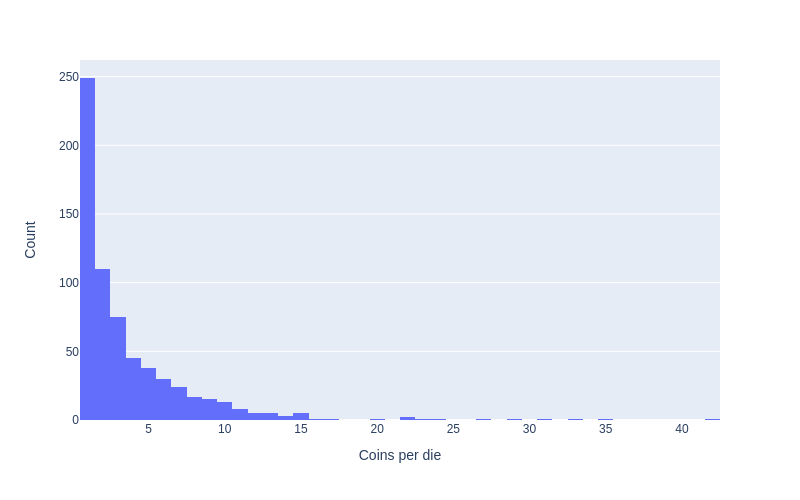}
    \caption{Distribution of number of coins per die, Paphos}
    \label{fig:paphos-distribution}
\end{figure}

Since only a small proportion of coins from Antiquity have been found to this day, coins follow uneven distributions across dies on both collections (Figure \ref{fig:paphos-distribution}). For example, 249 out of 2484 coins from the Paphos collection (9.98\%) are singletons - the only ones to correspond to their respective die.
Also note that we only consider the obverse of each coin (not the reverse). For manufacturing reasons (mostly for Greek coins), the reverse die used to break more often, leading to 1856 (74.7\%) singletons on the Paphos collection.

\subsubsection{Metrics}

We aim to find the partition of a collection separating coins upon dies. To evaluate the results, we use three standard external clustering validity measures, that evaluate the similarity between two partitions $U$ and $V$. All those metrics range from zero to one and are maximized for similar partitions.

The Adjusted Rand Index ($ARI$) \cite{vinh2009information} is based on the Rand Index ($RI$) that evaluates the proportion of sample pairs that are correctly partitioned, adjusted for chance against unbalanced clusters.

$$ARI = \frac{RI - E[RI]}{\text{max}(RI) - E[RI]}$$

Similarly, the Adjusted Mutual Information ($AMI$) \cite{meilua2007comparing} is an adjusted version of the Mutual Information ($MI$) in an Information Theory sense. In the general case, $AMI$ and $ARI$ are often close, but $AMI$ is supposed to be more relevant when the ground truth is unbalanced \cite{romano2016adjusting}. Noting $H(.)$ the Shannon Entropy, the $AMI$ of two partitions $U$ and $V$ is defined by :

$$AMI(U, V) = \frac{MI - E[MI]}{\text{mean}(H(U), H(V)) - E[MI]}$$

Lastly, the Fowlkes-Mallows Index ($FMI$) \cite{fowlkes1983method} is the geometric mean of pairwise Precision and Recall, and is generally more relevant with noisy data or unrelated partitions.
We also report pairwise Precision and Recall, since a significant difference between both could help explaining poor results in some cases.

\subsection{Performance on Die Studies}
\label{sec:result_perf}

% \todo[inline]{
% - Evoque reproductibility issue Heinecke? In practice, NMI presented in his paper isn't that impressive \\
% }

We present a comparison of our model with the fully automatic CADS model~\cite{taylor2020computer} on the datasets Paphos and Tanis (\autoref{tab:final-paphos-tanis}). We first observe that, in the CADS approach, the HDBSCAN clustering performs better than the best obtainable result with the agglomerative nesting algorithm that was used in \cite{taylor2020computer}. We also note that our method clearly outperforms CADS over every evaluation measure on both datasets, with most metrics above 0.90.

While the performances are much higher than CADS, some differences remain in comparison to die studies conducted by humans. The fraction of dies that are correctly identified among all those that should have been identified is reflected by the recall. The fraction of dies correctly identified among all those that have been identified is reflected by the precision. Since the number of dies is usually quite small in practice, a manual inspection of the identified groups can be conducted, leading to improved precision. However, the errors in terms of recall remain problematic since fixing them requires to inspect the full collection, thus losing the benefits of automation.

\begin{table}[!ht]
\caption{Die study results on the Paphos collection}
\label{tab:final-paphos-tanis}
\centering
\begin{tabular}{llllll|ccccc}
     \toprule
              & \multicolumn{5}{c|}{Paphos} & \multicolumn{5}{c}{Tanis}\\
     Approach & AMI & ARI & FMI & Prec. & Rec. & AMI & ARI & FMI & Prec. & Rec. \\
     \midrule
     {\small CADS-HDBSCAN} & 0.747 & 0.678 & 0.703 & 0.914 & 0.540 & 0.532 & 0.437 & 0.471 & 0.674 & 0.329 \\
     $CADS-AG^*$ & 0.664 & 0.483 & 0.517 & 0.742 & 0.360  & 0.499 & 0.382 & 0.445 & 0.769 & 0.258  \\
     \textbf{Ours} & \textBF{0.981} & \textBF{0.978} & \textBF{0.978} & \textBF{0.974} & \textBF{0.981} & \textBF{0.920} & \textBF{0.908} & \textBF{0.909} & \textBF{0.932} & \textBF{0.888}  \\
     \bottomrule
\end{tabular}
% \begin{tabular}{@{}llllll@{}}
%      \toprule
%      Approach & AMI & ARI & FMI & Precision & Recall \\
%      \midrule
%      $CADS-HDBSCAN$ & 0.747 & 0.678 & 0.703 & 0.914 & 0.540  \\
%      $CADS-AG^*$ & 0.664 & 0.483 & 0.517 & 0.742 & 0.360\\
%      \textbf{Ours} & \textBF{0.981} & \textBF{0.978} & \textBF{0.978} & \textBF{0.974} & \textBF{0.981}  \\
%      \bottomrule
% \end{tabular}
% %%% pour gain de place (FIXME pour camera ready)
% %\end{table}
% \vspace{0.5cm}
% %\begin{table}[!ht]
% \caption{Die study results on the Tanis 1986 collection}
% \label{tab:final-tanis}
% \centering
% \begin{tabular}{@{}llllll@{}}
%      \toprule
%      Approach & AMI & ARI & FMI & Precision & Recall \\
%      \midrule
%      $CADS-HDBSCAN$ & 0.532 & 0.437 & 0.471 & 0.674 & 0.329  \\
%      $CADS-AG^*$ & 0.499 & 0.382 & 0.445 & 0.769 & 0.258  \\
%      \textbf{Ours} & \textBF{0.920} & \textBF{0.908} & \textBF{0.909} & \textBF{0.932} & \textBF{0.888}  \\
%      \bottomrule
% \end{tabular}
\end{table}

\subsection{Ablation and Analysis}

%\subsubsection{Building a graph from matches and for clustering}
\subsubsection{On the use of Silhouette to find the optimal threshold}

%Label propagation clustering requires considering each pair of coins as either connected or disconnected. 
%For the moment, we have a scalar corresponding to the number of matches for each pair. Therefore, finding a threshold that correctly separates connected and disconnected edges would allow us to perform graph clustering.
%To do so, we binarize every edge upon a threshold determined via the Silhouette coefficient. The objective is then to evaluate the coherence of this approach.

As outlined in Section~\ref{sec:clustering}, we use the Silhouette Coefficient to establish a threshold for the number of matches, which allows us to transform coin similarities into a graph, on which we apply the Label Propagation algorithm. In this section, we aim to experimentally validate the relevance of this choice.

\begin{figure}[!ht]
    \centering
    \includegraphics[scale=0.4]{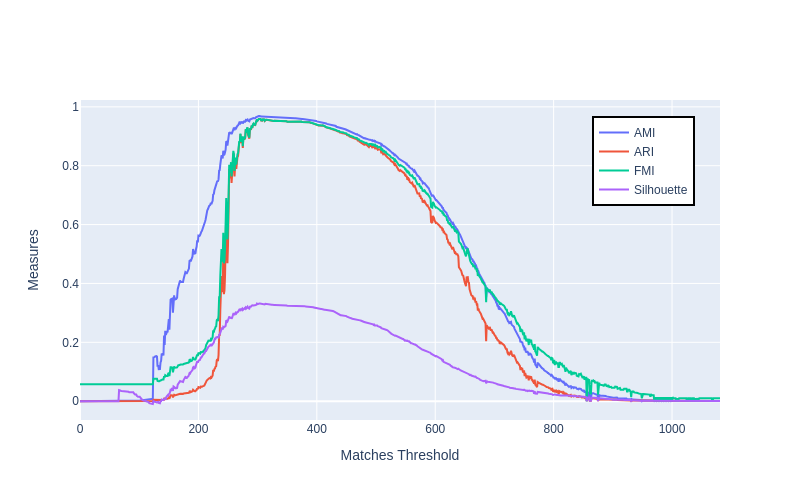}
    \caption{Quality of Die study from matches threshold on the Paphos dataset. We observe \textit{a posteriori} that the Silhouette Coefficient (without ground truth) is reliable to identify a threshold which leads to good results according to several metrics (ARI, AMI and FMI reflect the quality of the partition in comparison to the ground truth).}
    \label{fig:clustering-threshold}
\end{figure}

To achieve this, we compute partitions for every possible threshold and plot their Adjusted Mutual Information (AMI), Adjusted Rand Index (ARI), and Fowlkes-Mallows Index (FMI). These metrics help us evaluate the quality of the resulting clusters compared to the ground truth. Additionally, we assess the internal coherence of the clusters using the Silhouette Coefficient. The results of this study on the Paphos dataset are presented in \autoref{fig:clustering-threshold}.

We observe that all the metrics have very similar variations through different thresholds and are correlated to the Silhouette Coefficient. More precisely, if we compute the Pearson correlations of AMI, ARI and FMI with the Silhouette Coefficient, we obtain respective correlations of $0.973$, $0.951$ and $0.952$ and the threshold that maximizes the Silhouette also maximizes all the metrics. These experimental results are also verified on the Tanis dataset and confirm the relevance of the use of the Silhouette Coefficient to estimate automatically the best threshold used to create the graph.

%In fact, all the evaluation metrics are maximized for the same threshold on both Paphos and Tanis collections. 
%As the Silhouette Coefficient is an internal validation measure that does not require knowing the optimal partition, we should then be able to approximate the optimal threshold (with respect to AMI, ARI and FMI) with great precision on unknown collections.

%\todo[inline]{
%- Final Table Tanis (compare with Taylor-like distances), AMI / ARI / FMI  + Pairwise Precision / Recall \\
%- Final Table Paphos (same) \\
%- Hyperparameter sensitivity: $top_k$ in terms of External indices (AMI, ...)\\
%- Evoque computation time? As time is a limiting factor for man RoMa instead of XFeaual die studies
%}

%\paragraph{Ablation / Other tests}
%\todo[inline]{
%- Compare different clustering algorithms on the last stage (like 1st Exp on Taylor-like distance)\\
%- Heat Kernel distance : AMI maximized for very high $\tau$, outputing same partition as mismatches distance \\
%- Clustering from UMAP (projection in a euclidean space) ? \\
%- Using RoMa instead of XFeat* \\
%- Using different top-k maximum regions to match (4096 / 500 / 50, could test more values - Could even script it after XFeat* refactoring) \\
%}

\subsubsection{Comparison with other image matching models}

To extract the matches, we chose the XFeat model for its recognized efficiency and effectiveness. In this section, we compare this model with the RoMa model \cite{edstedt2024roma}, which obtains state-of-the-art results in numerous benchmarks \cite{li2018megadepth, mishkin2015wxbs} with the objective of estimating the fundamental matrix between two images. This model estimates a dense warp between two images of the same object and outputs pixel-level probabilities for its dense warp. We consider a match above a certain threshold $t$.

\begin{table}[!ht]
\caption{Clustering upon RoMa matches, Tanis collection}
\label{tab:roma-tanis}
    \centering
    \begin{tabular}{@{}llllll@{}}
        \toprule
        Approach & AMI & ARI & FMI & Precision & Recall \\
        \midrule
        XFeat & \textbf{0.920} & \textbf{0.908} & \textbf{0.909} & 0.932 & \textbf{0.888} \\
        %No Filter & - & - & - & - & - \\
        %HDBSCAN clustering & 0.917 & 0.900 & 0.902 & 0.897 & 0.907 \\
        RoMa, $t=0.8$ & 0.727 & 0.530 & 0.563 & 0.426 & 0.743 \\
        RoMa, $t=0.9$ & 0.745 & 0.699 & 0.709 & 0.793 & 0.634 \\ 
        RoMa, $t=0.95$ & 0.742 & 0.702 & 0.724 & 0.905 & 0.580 \\ % use 0.95 threshold
        RoMa, $t=0.99$ & 0.539 & 0.506 & 0.584 & \textbf{0.993} & 0.344 \\
        \bottomrule
    \end{tabular}
\end{table}

From the results presented in \autoref{tab:roma-tanis}, using RoMa matches instead of XFeat causes severe clustering degradations for every tested confidence threshold $t$. Especially with $t \leq 0.9$, pairwise recall is much lower than with XFeat, meaning that lots of coins from the same die are not considered connected in this case. With $t=0.8$, more than half of connected coins are from different dies. These results show that even though RoMa is state-of-the-art on \textit{SfM} tasks, it is outperformed by XFeat for the task considered in this study, which consists in quantifying matches between 2 different instances. Lastly, RoMa is much more computationally expensive, thus longer to compute, taking 22 hours to compute with two GeForce GTX TITAN X GPUs on the Tanis collection (295 coins), while it took only 5 minutes with XFeat. Therefore, we do not report results obtained with RoMa on the Paphos collection since it would take weeks of computation, which would be as long as doing the work manually. Our (perfectible) implementation of CADS required 2 days of calculations for Paphos and around 5 hours with our approach.
%Tanis:  5min13s for XFeat matches.

%\todo[inline]{Si RoMa mettait 22 heures sur Tanis avec 200 pièces, il aura tout juste fini sur Paphos avant la fin du stage...}

\subsubsection{Comparison with other clustering algorithms}
%\todo[inline]{Bouger ici les paragraphes sur le clustering avec HDBSCAN / UMAP+HDBSCAN? \\
%- Présenter ça comme "on teste notre méthode contre d'autres" ou "on en teste plusieurs pour sélectionner la meilleure"?}

%We test our clustering strategy using graph clustering with label propagation against other possibilities. We divide those possibilities in three groups : other graph clustering techniques, techniques clustering from a dissimilarity matrix, and techniques clustering from coordinates in a euclidean space.
To cluster coins from a similarity matrix of matches, our proposed method uses graph clustering, namely Label Propagation. In this section, we test our approach against other graph clustering algorithms and against other clustering approaches, directly based on the dissimilarity matrix or after projecting data in a Euclidean Space.

\paragraph{Graph Clustering} 
%\todo[inline]{
%- add compute time in table? or just enumerate exec time for 3 graph clustering and say "label prop is faster, hence preferable choice"?
%}
Concerning graph clustering algorithms, we test using the Louvain \cite{blondel2008louvain} and Leiden \cite{traag2019leiden} algorithms. These methods aim at maximizing the graph modularity, hence we use the same method to build a graph from a similarity before clustering. We also test using the graph's connected components as cluster labels (noted ``Connected'' in \autoref{tab:clustering-strat-paphos-tanis}).

\paragraph{Clustering from a dissimilarity matrix}

Given $M$ a matrix of pairwise matches between coins, we derive a dissimilarity matrix $D = max(M) - M$ and perform hierarchical clustering upon $D$ with HDBSCAN \cite{mcinnes2017hdbscan}. It builds a hierarchy between samples from their density without making any assumption on the structure of clusters. Furthermore, it is able to output uneven partitions with outliers, as die studies usually require. We use HDBSCAN with a minimum cluster size of 2 and use outliers as singletons. Output clusters therefore  have no minimum size.

\paragraph{Projection to a Euclidean Space}

Another possibility is to project our coins into a Euclidean Space before performing clustering. We use UMAP \cite{mcinnes2018umap}, a manifold learning technique suited for clustering \cite{allaoui2020considerably} that aims at preserving local more than global structure of data. From the dissimilarity matrix $D$, it projects data into $n$ dimensions, with $n$ a hyperparameter. Then, we perform clustering with HDBSCAN over the projected coordinates to obtain a partition. In a similar fashion as the method used to build a graph (\autoref{sec:clustering}), the optimal number of dimensions $n^*$ is determined using the Silhouette Coefficient of the partitions corresponding to every $n$ in $[2, 100]$. Since UMAP is not fully deterministic, it could lead to variations between different iterations. We thus report the average of 5 runs in ~\autoref{tab:clustering-strat-paphos-tanis}.

\begin{table}[!ht]
\caption{Results with different clustering strategies}
\label{tab:clustering-strat-paphos-tanis}
    \centering
    \begin{tabular}{@{}ll|lll|lll@{}}
        \toprule
         \multicolumn{2}{c|}{Clustering} & \multicolumn{3}{c|}{Paphos} & \multicolumn{3}{c}{Tanis} \\
        Input & Algorithm & AMI & ARI & FMI & AMI & ARI & FMI \\
        \midrule
        Graph & Label Propagation & \textbf{0.981} & \textbf{0.978} & \textbf{0.978} & \textbf{0.920} & \textbf{0.908} & \textbf{0.909}\\
        Graph & Louvain & \textbf{0.981} & \textbf{0.978} & \textbf{0.978}& \textbf{0.920} & \textbf{0.908} & \textbf{0.909}\\
        Graph & Leiden & \textbf{0.981} & \textbf{0.978} & \textbf{0.978}& \textbf{0.920} & \textbf{0.908} & \textbf{0.909}\\
        Graph & Connected & \textbf{0.981} & \textbf{0.978} & \textbf{0.978} & 0.915 & 0.903 & 0.905\\
        %\midrule
        Dissimilarity & HDBSCAN & 0.966 & 0.967 & 0.967 & 0.917 & 0.900 & 0.902\\
        %\midrule
        Projection & HDBSCAN & 0.882 & 0.830 & 0.837& 0.888 & 0.857 & 0.860\\
        \bottomrule
    \end{tabular}
\end{table}

%\todo[inline]{discuss clustering results : 
%- 3 graph clustering obtain very similar results, select the faster one : LabelProp }

A first observation can be made that the method using UMAP projection before HDBSCAN clustering in a Euclidean Space obtains the poorest clustering performances. Even though UMAP is supposedly suited for clustering, it makes an assumption on the ``number of neighbors'' a sample has. This approach is the one that is the less adapted to our use-case.
%\todo[inline]{On peut dire que UMAP considère 15 (reco dans le papier) points comme voisins, alors que rien ne nous prouve qu'une pièce a 15 autres pièces similaires.\\A quel point développer ça? Pas sûr que le but ce soit de ressortir les arguments de topologie de UMAP, pas le but du papier, surtout pour une xp peu concluante}

Then, clustering with HDBSCAN from a dissimilarity matrix obtains much higher results in every studied measure than projection methods. In particular, its large improvement upon the projection-based method assesses that clustering directly from a dissimilarity instead of using UMAP improves results for our case, contrary to the general case. Finally, graph-based techniques seem to offer the best clustering results on both datasets, with all of those methods reaching very similar performances on the Paphos collection. The high $AMI$, $ARI$ and $FMI$ obtained while using connected components as clusters labels seem to show that clustering performance issues more from graph construction than from graph clustering itself. This simple strategy obtains the highest results on Paphos. Yet, its sensitivity to erroneous connections as well as its lower results with Tanis should not make it a default choice.
On both collections, Label Propagation, Louvain and Leiden clustering obtain the same results, being the best solutions regarding $AMI$, $ARI$ and $FMI$. Note that Label Propagation is the fastest of those three approaches, typically being $1.7\times$ faster than Louvain, and $12.9\times$ than Leiden when computing every partition needed to determine the optimal graph threshold (\autoref{fig:clustering-threshold}).

%\todo[inline]{
%timeit results for graph clustering, every possible threshold for Tanis (0-1082)
%4.32, 7.4, 12.7
%}

\subsubsection{Impact of Filtering matches}

As not all matches from XFeat are kept, we can evaluate the importance of this filtering stage, both in terms of clustering performances on known datasets (\autoref{tab:results-filtering}) and in terms of hyperparameter stability, as it could be crucial on unknown datasets. 

\begin{figure}[!t]
    \centering
    \includegraphics[scale=0.4]{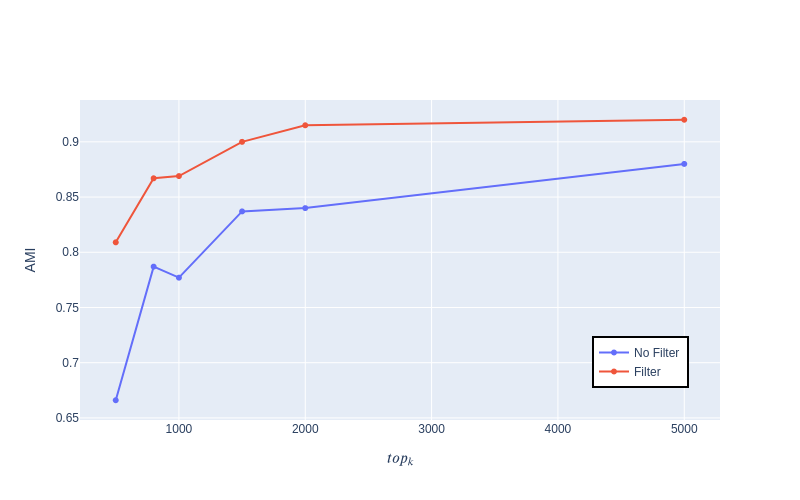}
    \caption{Variations of AMI on Tanis dataset upon the hyperparameter $top_k$ from Xfeat, with and without filtering}
    \label{fig:filtering-topk}
\end{figure}

The main hyperparameter used for computing XFeat matches is $top_k$, the maximum number of regions of the image to consider for matching (also note that a higher $top_k$ induces a higher computational cost). In  \autoref{fig:filtering-topk} we represent its importance towards AMI, with and without filtering. 

Two observations can be made from these results. First, filtering consistently increases clustering quality. Second, the AMI observes fewer variations through different $top_k$ values when filtering matches (the difference between higher and lower AMI on the studied range). Thus, filtering reduces the sensitivity to $top_k$. 
It can also be noted that $AMI$ stabilizes for $top_k > 5000$. For this reason we use a default $top_k=5000$ that should obtain optimal results. On a hypothetical unknown collection that would require $top_k > 5000$, filtering would greatly limit the negative impact of having a suboptimal value.

\begin{table}[!ht]
\caption{Impact of the filtering of the matches ($top_k = 5000$)}
\label{tab:results-filtering}
\centering
\begin{tabular}{l|lllll|lllll}
     \toprule
     \multirow{2}{*}{Approach} & \multicolumn{5}{c|}{Paphos dataset} & \multicolumn{5}{|c}{Tanis dataset} \\
      & AMI & ARI & FMI & Prec. & Recall & AMI & ARI & FMI & Prec. & Recall \\
     \midrule
     Filter & 0.981 & 0.978 & 0.978 & 0.975 & 0.981 & 0.920 & 0.908 & 0.909 & 0.932 & 0.888 \\
     No Filter & 0.973 & 0.968 & 0.968 & 0.985 & 0.952 & 0.880 & 0.854 & 0.861 & 0.950 & 0.780 \\
     \bottomrule
\end{tabular}
\end{table}

\subsubsection{Summary of Ablation Studies}

%\todo[inline]{Laisser tel quel? Si la suite sert principalement à développer l'importance de chaque point}
%\todo[inline]{
%- Rename "Summary of ablation studies"?\\
%- Laisser ligne pour RoMa dans le tableau recap de Paphos? Comme trop long pour le calculer
%}

To summarize the different experiments and ablations, \autoref{tab:ablation-paphos-tanis} enlightens the importance of the different aspects of our method.
The most severe degradation occurs when using RoMa matches instead of XFeat. 
%Furthermore, using MAGSAC++ to filter matches, and refine the similarity measure allows for noticeable improvements, especially towards pairwise recall. Those two observations show great importance in the extraction of the similarity matrix that is further used for clustering.
Clustering from a graph using Label Propagation allows the highest scores regarding every studied metric on both datasets.
Furthermore, using MAGSAC++ to filter matches has more impact towards clustering performances when clustering from a graph than from a dissimilarity matrix.

\begin{table}[!ht]
\caption{Ablation on Paphos and Tanis datasets, highlighting the influence of keypoints matching (Xfeat or RoMa), MAGSAC++ filtering, and the clustering approach we propose (AGLP) compared to HDBSCAN in various settings. Best results in bold and second best underlined. Computing RoMa on Paphos was too computationally expensive to be usable in practice.}
\label{tab:ablation-paphos-tanis}
\centering
\begin{tabular}{@{}lccc|lll|lll@{}}
     %\toprule
     %Approach & AMI & ARI & FMI & Precision & Recall \\
     %\midrule
     %Ours & 0.969 & 0.959 & 0.959 & 0.955 & 0.963  \\
     %Dissimilarity-HDBSCAN & 0.955 & 0.949 & 0.949 & 0.927 & 0.972  \\
     %Projection-HDBSCAN & 0.822 & 0.767 & 0.772 & 0.835 & 0.713 \\
     %No Filter & 0.962 & 0.953 & 0.953 & 0.962 & 0.944  \\
     %RoMa Matching & - &  - & - & - & -  \\
     %\bottomrule
    \toprule
     & \multirow{2}{*}{Matching} & \multirow{2}{*}{Filtering} & \multirow{2}{*}{Clustering}  & \multicolumn{3}{|c}{Paphos}& \multicolumn{3}{|c}{Tanis}\\
    \multicolumn{4}{c|}{} & AMI & ARI & FMI & AMI & ARI & FMI \\
    \midrule
    \textBF{ours} & \textBF{XFeat} & \textBF{yes} & \textBF{AGLP}  & \textBF{0.981} & \textBF{0.978} & \textBF{0.978}  & \textBF{0.920} & \textBF{0.908} & \textBF{0.909}\\
    \multirow{5}{*}{ \rotatebox{90}{baselines}} & XFeat & no & AGLP  & \underline{0.973} & \underline{0.968} & \underline{0.968} & 0.880 & 0.854 & 0.861\\
    & XFeat & yes & {\scriptsize $HDBSCAN_{dissim.}$} & 0.966 & 0.967 & 0.967 & \underline{0.917} & \underline{0.900} & \underline{0.902}\\%& 0.897 & 0.907\\
    & XFeat & no & {\scriptsize $HDBSCAN_{dissim.}$} & 0.967 & 0.964 & 0.967 & 0.901 & 0.885 & 0.887\\
    & XFeat & yes & {\scriptsize $HDBSCAN_{proj.}$} & 0.882 & 0.830 & 0.837 & 0.888 & 0.857 & 0.860\\%& 0.871 & 0.849\\
    & RoMa & no & AGLP & - & - & - & 0.742 & 0.702 & 0.724 \\
    \bottomrule
\end{tabular}
\end{table}

\begin{figure}[tb]
     \centering
     \begin{subfigure}[b]{0.45\textwidth}
         \centering
         \includegraphics[width=\textwidth]{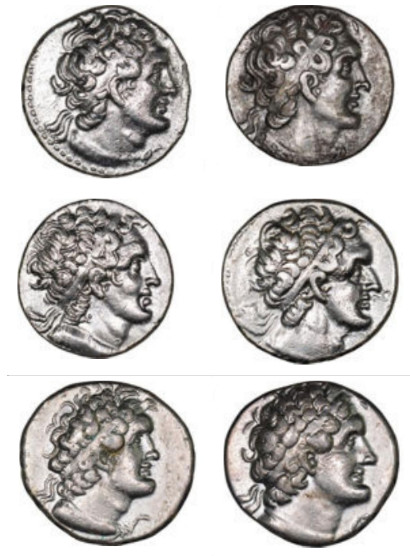}
         \caption{False Positives}
         \label{subfig:false-positive}
     \end{subfigure}
     \begin{subfigure}[b]{0.45\textwidth}
         \centering
         \includegraphics[width=\textwidth]{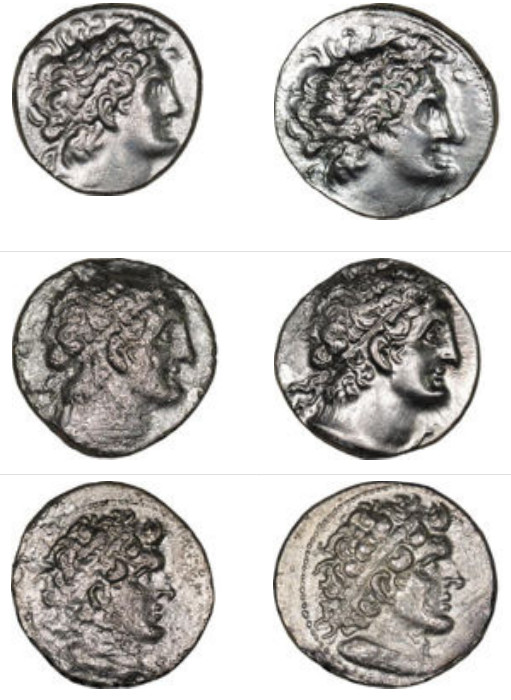}
         \caption{False Negatives}
         \label{subfig:false-negative}
     \end{subfigure}
    \caption{Example of errors with our method. Pairs in \ref{subfig:false-positive} are found struck by the same die by our approach while they are not. % have high similarities for pairs struck by different dies (top to bottom: 316, 288 and 194 matches). 
    Pairs in \ref{subfig:false-negative} are found struck by different dies when in fact the same die was used.}
    % Pairs in 6a have the highest computed similarities for pairs struck by different dies, while pairs in 6b have the lowest for coins struck with the same die
    \label{fig:largest-errors}
\end{figure}

\section{Conclusion and Limitations}
%conclusion and future works

We proposed a fully automated solution for performing die studies. We did this by using a fast and robust keypoint detector+descriptor %Image Matching techniques usually used in an $\textit{SfM}$ framework 
to find matches between every pair of coins. We then construct an optimal graph using intrinsic clustering metrics, and perform simple graph clustering to extract a partition upon dies. Our solution significantly improves on the state-of-the-art, and does not require any human intervention in its process. 
While the errors of our approach in terms of precision (less than 10\%) can be further fixed manually (see~\autoref{sec:result_perf}), those in terms of recall remain problematic since fixing them requires to inspect the full collection manually, thus losing the benefits of automation. For existing die studies, which may not be perfect, our approach may help numismatists to quickly refine the initial annotation by providing false positive and negative samples to be re-examined (\autoref{fig:largest-errors}).

% \todo[inline]{
% - bronze peut être plus difficile que argent (mais Or plus facile) \\
% }
Thanks to this automated approach which code is released, %\footnote{released at https://anonymous\_for\_review after acceptance and before publication}, 
ancient historians will be able to carry out die studies on a much larger scale, and thus deepen our understanding of the ancient world. This will be particularly useful in the case of civilizations for which die studies could not be completely carried out due to the size of the available corpora, such as Athen, Aegina or Roman Empire. %It nevertheless remains limited to the civilizations with sufficient monetary treasures. Moreover
This may also make it possible to study collections from several coin workshops and make links that humans would have thought impossible or unlikely a priori. 
However, the two corpora considered in this paper can be considered as quite homogeneous, since all the coins are represented by RGB pictures with about the same size, illumination conditions, viewpoint and resolution. Other collections likely to be the subjects of die studies can be much more heterogeneous when the various coins have been captured by different historians (or individuals), at different times, with different photo equipment. In some cases, all we have is a scan of the negative of the original photo, or the one printed in a historical research article, or even the image of a (plaster) mold of the coins or dies. The proposed approach is likely to be much less effective in such cases, in particular because the keypoints detector currently used is adapted to natural RGB images. Matching such heterogeneous data is an ambitious challenge, but is necessary to address die studies in all their diversity.

\noindent
\textbf{Acknowledgment} this work was partially funded by the Agence Nationale de la Recherche (ANR) for the STUDIES project  ANR-23-CE38-0014-02.
%Cette recherche a été financée en tout ou partie, par l’Agence Nationale de la Recherche (ANR) au titre du projet « ANR-23-CE38-0014-02 »

\bibliographystyle{splncs04}
\bibliography{main}
\end{document}